\def\year{2020}\relax
\documentclass[letterpaper]{article} 
\usepackage{aaai20}  
\usepackage{times}  
\usepackage{helvet} 
\usepackage{courier}  
\usepackage[hyphens]{url}  
\usepackage{graphicx} 

\usepackage{amsmath}
\usepackage{multirow}
\usepackage{setspace}
\usepackage{txfonts}

\usepackage{graphicx}  
\title{Hybrid Graph Neural Networks for Crowd Counting}
\author{Ao Luo\textsuperscript{\rm 1}, Fan Yang\textsuperscript{\rm 2}, Xin Li\textsuperscript{\rm 2}, Dong Nie\textsuperscript{\rm 3}, Zhicheng Jiao\textsuperscript{\rm 4}, Shangchen Zhou\textsuperscript{\rm 5}, Hong Cheng\textsuperscript{\rm 1}\thanks{denotes the Corresponding Author}\\
\textsuperscript{\rm 1}Center for Robotics, University of Electronic Science and Technology of China\\
\textsuperscript{\rm 2}Inception Institute of Artificial Intelligence\\
\textsuperscript{\rm 3}Department of Computer Science, University of North Carolina at Chapel Hill\\
\textsuperscript{\rm 4}University of Pennsylvania \quad\quad\quad \textsuperscript{\rm 5}Nanyang Technological University\\
aoluo\_uestc@hotmail.com, hcheng@uestc.edu.cn  
}
 
\begin{document}

\maketitle

\begin{abstract}
Crowd counting is an important yet challenging task due to the large scale and density variation. Recent investigations have shown that distilling rich relations among multi-scale features and exploiting useful information from the auxiliary task, i.e., localization, are vital for this task. Nevertheless,  how to comprehensively leverage these relations within a unified network architecture is still a challenging problem. In this paper, we present a novel network structure called Hybrid Graph Neural Network ({\scshape{HyGnn}}) which targets to relieve the problem by interweaving the multi-scale features for crowd density as well as its auxiliary task (localization) together and performing joint reasoning over a graph. Specifically, {\scshape{HyGnn}} integrates a hybrid graph to jointly represent the task-specific feature maps of different scales as nodes, and two types of relations as edges: {\bfseries (i)} multi-scale relations for capturing the feature dependencies across scales and {\bfseries (ii)} mutual beneficial relations building bridges for the cooperation between counting and localization. Thus, through message passing, {\scshape{HyGnn}} can distill rich relations between the nodes to obtain more powerful representations, leading to robust and accurate results. Our {\scshape{HyGnn}} performs significantly well on four challenging datasets: ShanghaiTech Part A, ShanghaiTech Part B, UCF\_CC\_50 and UCF\_QNRF, outperforming the state-of-the-art approaches by a large margin. 
\end{abstract}

\section{Introduction}

Crowd counting, with the purpose of analyzing large crowds quickly, is a crucial yet challenging computer vision and AI task. It has drawn a lot of attention due to its potential applications in public security and planning,  traffic control, crowd management, public space design, \emph{etc.}

Same as many other computer vision tasks, the performance of crowd counting has been substantially improved by Convolutional Neural Networks (CNNs). Recently, the state-of-the-art crowd counting methods~\cite{Liu_2019_CVPR,Liu1_2019_CVPR,Wan_2019_CVPR,Liu1_2019_CVPR,Jiang_2019_CVPR}  mostly follow the \emph{density-based} paradigm. Given an image or video frame, CNN-based regressors are trained to estimate the crowd density map, whose values are summed to give the entire crowd count. 

\begin{figure}[pt]
	\begin{center}
		\includegraphics[width=0.96\linewidth]{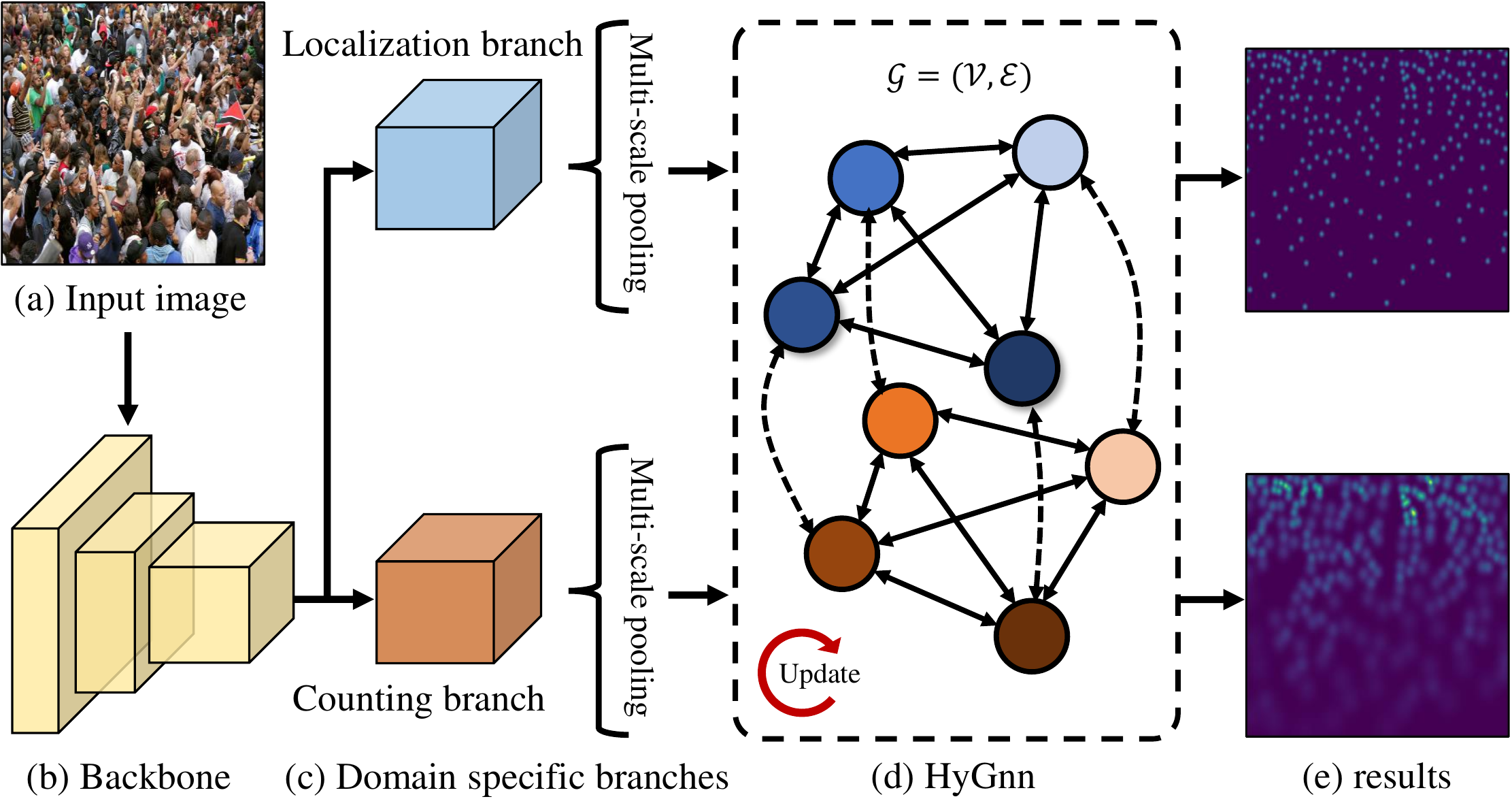}
	\end{center}
	\caption{Illustration of the proposed {\scshape{HyGnn}} model. (a) Input image, in which crowds have heavy overlaps and occlusions. (b) Backbone, which is a truncated VGG-16 model. (c) Domain-specific branches: one for crowd counting and the other for localization. (d){\scshape{HyGnn}}, which represents the features from different scales and domains as nodes, while the relations between them as edges. After several message passing iterations, multiple types of useful relations are built. (e) Crowd density map (for counting) and localization map (as the auxiliary task).}
	\label{fig:1}
\end{figure}

Recent studies~\cite{Shen_2018_CVPR,Cao_2018_ECCV,li2017multi,li2018contour} have shown that multi-scale information, or \emph{relations} across multiple scales helps to capture contextual knowledge which benefits crowd counting.  Moreover, the crowd counting and its auxiliary task (localization), in spite of analyzing the crowd scene from different perspectives, could provide beneficial clues for each other~\cite{Liu_2019_CVPR,Lian_2019_CVPR}. Crowd density map can offer guidance information and self-adaptive perception for precise crowd localization, and on the other hand, crowd localization can help to alleviate local inconsistency issue in density map. The mutual cooperation, or called \emph{mutual beneficial relation}, is the key factor in estimating the high-quality density map. However, most methods only consider the crowd counting problem from one aspect, while ignore the other one. Consequently, they fail to fully utilize multiple types of useful relations or structural dependencies in the learning and inferring processes, resulting in sub-optimal results.

One primary reason is the lack of a unified and effective framework capable of modeling the different types of relations (\emph{i.e.,} multi-scale relations and mutual beneficial relations) over a single model.  To address this issue, we introduce a novel Hybrid Graph Neural Network ({\scshape{HyGnn}}), which formulates the crowd counting and localization as a graph-based, joint reasoning procedure. As shown in Fig.~\ref{fig:1}, we build a hybrid graph which consists of two types of nodes, \emph{i.e.}, counting nodes storing density-related features and localization nodes storing location-related features. Besides, there are two different pairwise relationships (edge types) between them. By interweaving the multi-scale and multi-task features together and progressively propagating information over the hybrid graph, {\scshape{HyGnn}} can fully leverage the different types of useful information, and is capable of distilling the valuable, high-order relations among them for much more comprehensive crowd analysis.

{\scshape{HyGnn}}  is easy to implement and end-to-end learnable. Importantly, it has two major benefits in comparison to existing models for crowd counting~\cite{Liu_2019_CVPR,Liu1_2019_CVPR,Wan_2019_CVPR,Liu1_2019_CVPR,Jiang_2019_CVPR}. {\bfseries (i)} {\scshape{HyGnn}} interweaves crowd counting and localization with a joint, multi-scale and graph-based processing rather than a simple combination as done in most existing solutions. Thus, {\scshape{HyGnn}} significantly strengthens the information flow between tasks and across scales, thereby enabling the augmented representation to incorporate more useful priors learned from the auxiliary task and different scales. {\bfseries (ii)} {\scshape{HyGnn}} explicitly models and reasons all relations (multi-scale relations and mutual beneficial relations) simultaneously over a hybrid graph, while most existing methods are not capable of dealing with such complicated relations. Therefore, our {\scshape{HyGnn}} can effectively capture their dependencies to overcome inherent ambiguities in the crowd scenes. Consequently, our predicted crowd density map is potentially more accurate, and consistent with the true crowd localization. 

In our experiments, we show that {\scshape{HyGnn}} performs remarkably well on four well-used benchmarks and surpasses prior methods by a large margin. Our {\bfseries contributions} are summarized in three aspects:

\begin{itemize}
	\item We present a novel end-to-end learnable model, namely Hybrid Graph Neural Network ({\scshape{HyGnn}}),  for joint crowd counting and localization. To the best of our knowledge, {\scshape{HyGnn}} is the first deep model capable of explicitly modeling and mining high-level relations between counting and its auxiliary task (localization) across different scales through a hybrid graph model. 
	
	\item {\scshape{HyGnn}} is equipped with a unique multi-tasking property, where different types of nodes, connections (or edges), and message passing functions are parameterized by different neural designs. With such property,  {\scshape{HyGnn}} can more precisely leverage cooperative information between crowd counting and localization to boost the counting performance.

	\item We conduct extensive experiments on four well-known benchmarks including ShanghaiTech Part A, ShanghaiTech Part B, UCF\_CC\_50 and UCF\_QNRF, on which we set new records.

\end{itemize}    

\begin{figure*}[pt]
	\begin{center}
		\includegraphics[width=0.9\linewidth]{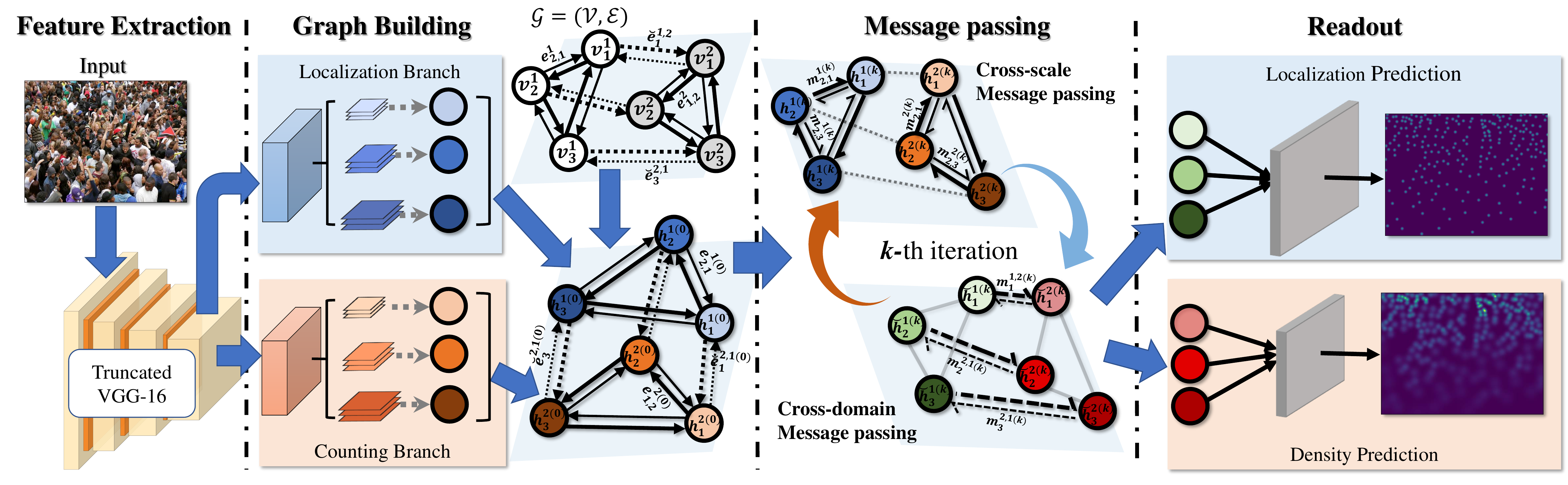}
	\end{center}
	\caption{Overall of our {\scshape{HyGnn}} model. Our model is built on the truncated VGG-16, and includes a Domain-specific Feature Learning Module to extract features from different domains. A novel {\scshape{HyGnn}} is used to distill multi-scale and cross-domain information, so as to learn better  representations. Finally, the multi-scale features are fused to produce the density map for counting as well as generate the auxiliary task prediction (localization map).}
	\label{fig:2}
\end{figure*}

\section{Related Works}

\subsubsection{Crowd Counting and Localization.} 

Early works~\cite{viola2005detecting1} in crowd counting use \emph{detection-based} methods, and employ handcrafted features like Haar~\cite{viola2001rapid} and HOG~\cite{dalal2005histograms} to train the detector. The overall performances of these algorithms are rather limited due to various occlusion. The \emph{regression-based} methods, which can avoid solving the hard detection problem, has become mainstream and achieved great performance breakthroughs. Traditionally, the regression models~\cite{chen2013cumulative,lempitsky2010learning,pham2015count} learn the mapping between low-level images features and object count or density, using gaussian process or random forests regressors. Recently, various CNN-based counting methods have been proposed~\cite{zhang2015cross,zhang2016single,Liu_2019_CVPR,Liu1_2019_CVPR,Wan_2019_CVPR,Liu1_2019_CVPR,Jiang_2019_CVPR} to better deal with different challenges, by predicting a density map whose values are summed to give the count. Particularly, the scale variation issue has attracted the most attention of recent CNN-based methods~\cite{DSN,SAAN}. On the other hand, as observed by some recent researches~\cite{Idrees_2018_ECCV,Liu_2019_CVPR,Lian_2019_CVPR}, although the current state-of-the-art methods can report accurate crowd count, they may produce the density map which is inconsistent with the true density. One major reason is the lack of crowd localization information. Some recent studies~\cite{Zhao_2019_CVPR,Liu_2019_CVPR}  have tried to exploit the useful information from localization in a unified framework. They, however, only simply share the underlying representations or interweave two modules for different task together for more robust representations. Differently, our {\scshape{HyGnn}} considers a better way to utilize the mutual guidance information: explicitly modeling and iteratively distilling the mutual beneficial relations across scales within a hybrid graph. For a more comprehensive survey, we refer interested readers to~\cite{kang2018beyond}.

\subsubsection{Graph Neural Networks.} 

The essential idea of Graph Neural Network (GNN) is to enhance the node representations by propagating information between nodes. Scarselli \emph{et al.}~\cite{scarselli2008graph} first introduced the concept of GNN, which extended recursive neural networks for processing graph structure data. Li \emph{et al.}~\cite{li2015gated} proposed to improve the representation capacity	of GNN by using Gated Recurrent Units (GRUs). Gilmer \emph{et al.}~\cite{GilmerSRVD17} used message passing neural network to generalize the GNN. Recently, GNN has been successfully applied in attributes recognition~\cite{Meng_2018_ECCV}, human-object interactions~\cite{Qi_2018_ECCV}, action recognition~\cite{Si_2018_ECCV}, \emph{etc.} Our {\scshape{HyGnn}} shares similar ideas with above methods that fully exploits the underlying relationships between multiple latent representations through GNN. However, most existing GNN-based models are designed to deal with \emph{only one} relation type, which may limit the power of GNN. To overcome above limitation, our {\scshape{HyGnn}} is equipped with a \emph{multitasking} property, \emph{i.e.,} parameterizing different types of connections (or edges) and the message passing functions with different neural designs,  which significantly discriminates {\scshape{HyGnn}} from all existing GNNs.

\section{Methodology}
\subsection{Preliminaries}

\subsubsection{Problem Formulation.} 

Let the crowd counting model be represented by the function $\mathcal{M}$ which takes an Image $\mathcal{I}$ as input and generates the corresponding crowd density map $\mathcal{D}$ (for counting) as well as the auxiliary task prediction, \emph{i.e.}, localization map $\mathcal{L}$.  Let $\mathcal{D}^g$ and $\mathcal{L}^g$ be the groundtruth of density map and localization map, respectively. Our goal is to learn the powerful \emph{domain-specific} representations, denoted as $\mathbf{f}_{d}$ and $\mathbf{f}_{l}$, to minimize errors between the estimated $\mathcal{D}$ and groundtruth $\mathcal{D}^g$, as well as $\mathcal{L}$ and $\mathcal{L}^g$. Notably, the two tasks share a common \emph{meta-objective}, and $\mathcal{D}^g$ and $\mathcal{L}^g$ are obtained from the same point-annotations without additional target labels.

\subsubsection{Notations.} 

To achieve the goal, we need to distill the underlying dependencies between multi-task and multi-scale features. Given the multi-scale density feature maps $\mathcal{F}_{d}=\{\mathbf{f}_d^{s_i}\}_{i=1}^N$ and multi-scale localization feature maps $\mathcal{F}_{l}=\{\mathbf{f}_l^{s_i}\}_{i=1}^N$, we represent $\mathcal{F}_{d}$ and $\mathcal{F}_{l}$ with a directed graph $\mathcal{G} = (\mathcal{V,E})$, where $\mathcal{V}$ is a set of nodes and $\mathcal{E}$ are edges. The nodes in our {\scshape{HyGnn}} are further grouped into two types: $\mathcal{V} = \mathcal{V}^1 \bigcup \mathcal{V}^2$, where $\mathcal{V}^1 = \{v^{1}_i\}_{i=1}^{|\mathcal{V}^1|}$ is the set of counting (density) nodes and $\mathcal{V}^2 = \{v^{2}_i\}_{i=1}^{|\mathcal{V}^2|}$ denotes the set of localization nodes. In our model, we have the same number of nodes in two latent domains, and therefore $|\mathcal{V}^1| = |\mathcal{V}^2| = N$. Accordingly, there are two types of edges $\mathcal{E} = \mathcal{E}_{i,j} \bigcup  \breve{\mathcal{E}}_{m,n}$ between them: {\bfseries (i)} cross-scale edges $e^m_{i,j} = (v^m_i, v^m_j) \in \mathcal{E}_{i,j}$ stand for the multi-scale relations between nodes from the $i_{th}$ scale to the $j_{th}$ scale within the same domain $m \in \{1, 2\}$, where $i,j \in \{1,\cdots,N\}$; {\bfseries (ii)} cross-domain edges $\breve{e}^{m,n}_{i}=({v}^{m}_i, {v}^{n}_i) \in \breve{\mathcal{E}}_{m,n}$ reflect mutual beneficial relations between nodes from the domain $m$ to $n$ with the same scale $i \in \{1,\cdots,N\}$, where $m,n \in \{1,2\}$ \& $m\ne n$. For each node ${v}^{m}_i$ ($i \in \{1,\cdots,N\}$ \& $m \in \{1,2\}$), we learn its updated representation, namely $\mathbf{h}^{m}_i$, through aggregating representations of its neighbors. Finally, the updated multi-scale features $\mathcal{H}_1=\{\mathbf{h}^{1}_{i}\}_{i=1}^{|\mathcal{V}^1|}$ and $\mathcal{H}_2=\{\mathbf{h}^{2}_{i}\}_{i=1}^{|\mathcal{V}^2|}$ are fused to produce the final representation $\mathbf{f}_d$ and $\mathbf{f}_l$, which are used to generate the outputs $\mathcal{D}$ and $\mathcal{L}$. Here, we only consider multi-scale relations between nodes in the same domain, and mutual beneficial (cross-domain) relations between nodes with the same scale in our graph model. Considering that our graph model is designed to simultaneously deal with two different node and relation types, we term it as Hybrid Graph Neural Network ({\scshape{HyGnn}}) which will be detailed in the following section.

\subsection{Hybrid Graph Neural Network ({\scshape{HyGnn}})}

\subsubsection{Overview.}
The key idea of our {\scshape{HyGnn}} is to perform $K$ message propagation iterations over $\mathcal{G}$ to joint distill and reason all relations between crowd counting and the auxiliary task (localization) across scales. Generally, as shown in Fig.~\ref{fig:2}, {\scshape{HyGnn}} maps the given image $\mathcal{I}$ to the final predictions $\mathcal{D}$ and $\mathcal{L}$ through three phases. {\bf \small First}, in the domain-specific feature extracting phase, {\scshape{HyGnn}} generates the multi-scale density features $\mathcal{F}_{d}$ and localization features $\mathcal{F}_{l}$ for $\mathcal{I}$ through a Domain-specific Feature Learning Module (DFL), and represents these features with a graph $\mathcal{G} = (\mathcal{V,E})$. {\bf \small Second}, a parametric message passing phase runs for $K$ times to propagate the message between nodes and also to update the node representations according to the received messages within the graph $\mathcal{G}$. {\bf \small Third}, a readout phase fuses the updated multi-scale features $\mathcal{H}_{1}$ and $\mathcal{H}_{2}$ to generate final representations (\emph{i.e.}, $\mathbf{f}_d$ and $\mathbf{f}_l$), and maps them to the outputs $\mathcal{D}$ and $\mathcal{L}$. Note that, as crowd counting is our main task, we emphasize the accuracy of $\mathcal{D}$ during the learning process.

\subsubsection{Domain-specific Feature Learning Module (DFL).} DFL is one of the major modules of our model, which extracts the multi-scale, domain-specific features $\mathcal{F}_{d}=\{\mathbf{f}_d^{s_i}\}_{i=1}^{N}$ and $\mathcal{F}_{l}=\{\mathbf{f}_d^{s_i}\}_{i=1}^{N}$ from the input  $\mathcal{I}$. DFL is composed of three parts: one front-end and two domain-specific back-ends. 

The front-end $Fr(\cdot)$ is based on the well-known VGG-16, which maps the RGB image $\mathcal{I}$ to the shared underlying representations: $\mathbf{f}_{share} = Fr(\mathcal{I})$. More specifically, the first 10 layers of VGG-16 are deployed as the front-end which is shared by the two tasks. Meanwhile, two series of convolution layers with different dilation rates are appended onto the back-ends, denoted as $B_d(\cdot)$ and $B_l(\cdot)$. With the large receptive fields, the stacked convolutions are tailored for learning domain-specific features: $\mathbf{f}_{d} = B_d(\mathbf{f}_{share})$ and $\mathbf{f}_{l} = B_l(\mathbf{f}_{share})$. In addition, the Pyramid Pooling Module (PPM)~\cite{zhao2017pyramid} is applied in each domain-specific back-end for extracting multi-scale features, followed by an interpolation layer $R(\cdot)$ to ensure the multi-scale feature maps to have the same size ${H\times W}$. 

\begin{figure}[pt]
	\begin{center}
		\includegraphics[width=0.85\linewidth]{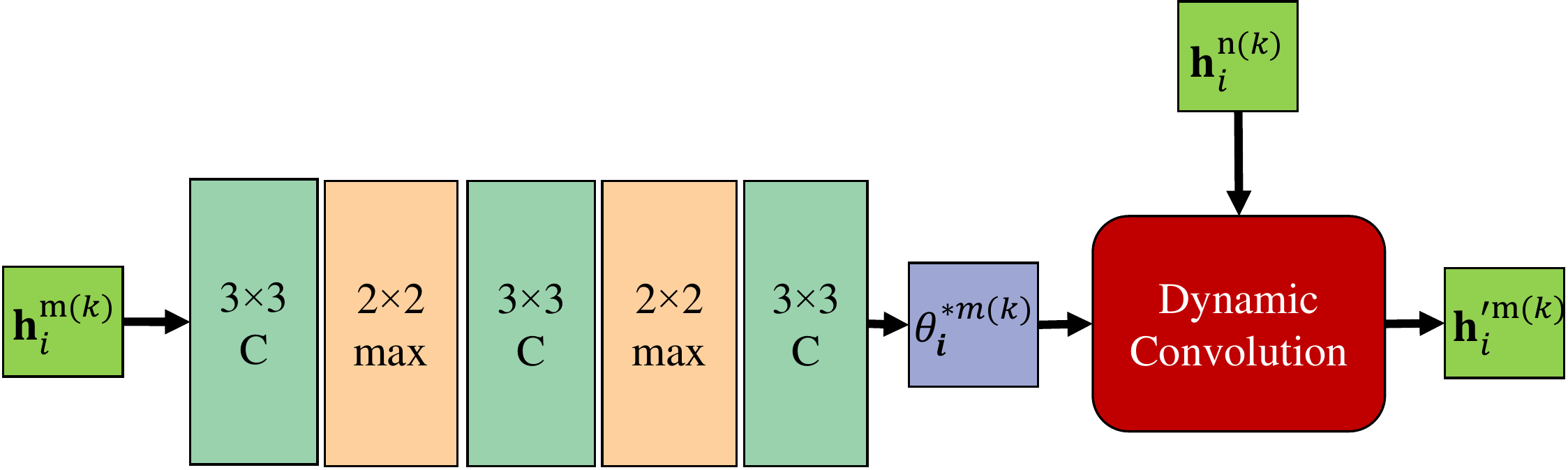}
	\end{center}
	\caption{The architecture of the learnable adapter. The adapter takes the node representation of one (source) domain $\mathbf{h}^{m(k)}_{i}$ as input and outputs the adaptive convolution parameters ${\theta^{*}}_i^{m(k)}$. The adaptive representation $\mathbf{h^\prime}^{m(k)}_{i}$ is generated conditioned on $\mathbf{h}^{n(k)}_{i}$.}
	\label{fig:4}
\end{figure}

\subsubsection{Node Embedding.} 
In our {\scshape{HyGnn}}, 
each node ${v}^{1}_i$ or ${v}^{2}_i \in \mathcal{V}$, where $i$ takes an unique value from $\{1, \cdots, |\mathcal{V}|\}$, is associated with an initial node embedding (or node state), namely $\mathbf{v}_i^1$ or $\mathbf{v}_i^2$. We use the domain-specific feature maps produced by DFL as the initial node representations. Taking an arbitrary counting node ${v}^{1}_i \in \mathcal{V}^1$ for example, its initial representations $\mathbf{h}^{1(0)}_i$ can be calculated by:
\begin{equation}
\mathbf{h}^{1(0)}_{i} = \mathbf{v}_{i}^1 = R(P(\mathbf{f}_{d}, s_i)) \in \mathbb{R}^{H\times W \times C},
\label{eq1}
\end{equation}
\noindent where $\mathbf{h}^{1(0)}_i\in \mathbb{R}^{H\times W \times C}$ is a 3D tensor feature (batch size is omitted). $R(\cdot)$ and $P(\cdot)$ denote the interpolation operation and pyramid pooling operation, respectively. The initial representation for the localization node ${v}^{2}_i \in \mathcal{V}^2$ is defined similarly as follows:
\begin{equation}
\mathbf{h}^{2(0)}_{i} = \mathbf{v}_{i}^2 = R(P(\mathbf{f}_{l}, s_i)) \in \mathbb{R}^{H\times W \times C},
\label{eq2}
\end{equation}
\noindent where $\mathbf{h}^{2(0)}_i\in \mathbb{R}^{H\times W \times C}$ denotes the initial representation for the localization node $v_i^2 \in \mathcal{V}^2$.

\subsubsection{Cross-scale Edge Embedding.} A cross-scale edge $e_{i,j}^m \in \mathcal{E}_{i,j}$ connects two nodes $v^{m}_i$ and $v^{m}_j$ which are from the same domain $m \in \{1,2\}$ but different scales $i,j \in \{1, \cdots,N\}$. The cross-scale edge embedding, denoted as $\mathbf{e}_{i,j}^m$, is used to distill the multi-scale relation from $v^{m}_i$ to $v^{m}_j$ as the edge representation. To this goal, we employ a relation function ${f_{rel}}(\cdot,\cdot)$ to capture the relations by:
\begin{equation}
\mathbf{e}_{i,j}^{m(k)} ={f_{rel}}(\mathbf{h}^{m(k)}_{i},\mathbf{h}^{m(k)}_{j}) = Conv(\mathrm{g} (\mathbf{h}^{m(k)}_{i},\mathbf{h}^{m(k)}_{j}))\in \mathbb{R}^{H\times W \times C},
\label{eq3}
\end{equation}

\noindent where $\mathrm{g}(\cdot,\cdot)$ is a function to combine the feature $\mathbf{h}^{m(k)}_{i}$ and $\mathbf{h}^{m(k)}_{j}$. Following~\cite{dgcnn}, we model $\mathrm{g}(\mathbf{h}_{i}, \mathbf{h}_{j}) = \mathbf{h}_{i} - \mathbf{h}_{j}$, making the relations based on the difference between node embeddings to alleviate the \emph{symmetric} impact in feature combination. $Conv(\cdot)$ means the convolution operation that is used to learn the edge embedding in a data-driven way. Each element in $\mathbf{e}_{i,j}^{m(k)}$ reflects the pixel-level relations between the nodes of different scales from $i$ to $j$. As a result, $\mathbf{e}_{i,j}^{m(k)}$ can be considered as the features that depict the \emph{multi-scale relationships} between nodes. 

\begin{figure}[pt]
	\begin{center}
		\includegraphics[width=0.8\linewidth]{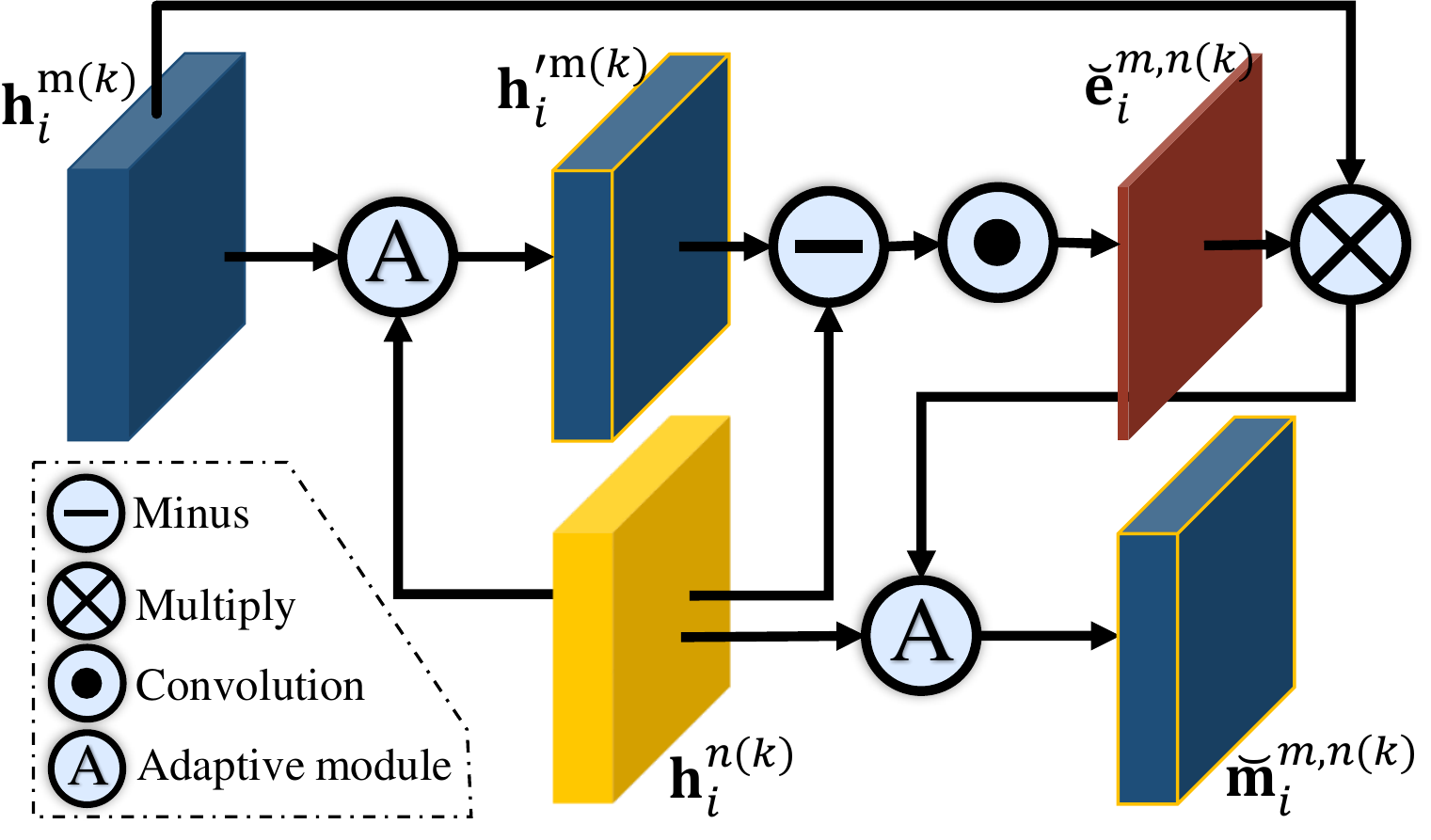}
	\end{center}
	\caption{Detailed illustration of the cross-domain edge embedding and message aggregation. Please see text for details.}
	\label{fig:3}
\end{figure}



\subsubsection{Cross-domain Edge Embedding.} Since our {\scshape{HyGnn}} is designed to fully exploit the complementary knowledge contained in the nodes of different domains ($m,n\in \{1,2\}$ \& $m\ne n$), one major challenge is to overcome the ``domain gap'' between them. Rather than directly combining features as used in the cross-scale edge embedding, we first adapt the node representation of one (source) domain $\mathbf{h}^{m(k)}_{i}$ conditioned on the node representation of the other (target) domain $\mathbf{h}^{n(k)}_{i}$ to overcome the domain difference. Here,  inspired by~\cite{bertinetto2016learning}, we integrate a learnable adapter $\mathcal{A}_m(\mathbf{h}^{m(k)}_{i} {\big|\big|} \mathbf{h}^{n(k)}_{i})$ into our {\scshape{HyGnn}} to transform the original node representation $\mathbf{h}^{m(k)}_{i}$ to the adaptive representation $\mathbf{h^\prime}^{m(k)}_{i}$ as follows:
\begin{equation}
\begin{aligned}
\mathbf{h^\prime}^{m(k)}_{i}&=\mathcal{A}_m(\mathbf{h}^{m(k)}_{i} {\big|\big|} \mathbf{h}^{n(k)}_{i})= {\theta^{*}}_i^{m(k)} * \mathbf{h}^{n(k)}_{i}, \\
& where \ {\theta^{*}}_i^{m(k)} = E_\phi(\mathbf{h}^{m(k)}_{i}).
\label{eq4}
\end{aligned}
\end{equation}
\noindent In the above function, $*$ is the convolution operation, and ${\theta^{*}}_i^{m(k)}$ means the dynamic convolutional kernels. $E_\phi(\cdot)$ is a one-shot learner to predict the dynamic parameters ${\theta^{*}}_i^{m(k)}$ from a single exemplar. Following~\cite{Nie_2018_CVPR}, as shown in Fig.~\ref{fig:4}, we implement it by a small CNN with learnable parameters $\phi$. 

After achieving the adaptive representation $\mathbf{h^\prime}^{m(k)}_{i}$, the cross-domain edge embedding $\breve{\mathbf{e}}^{m,n}_i$ for the edge $\breve{e}^{m,n}_{i}=({v}^{m}_i, {v}^{n}_i) \in \mathcal{E}_{m,n}$ can be formulated as:
\begin{equation}
\begin{aligned}
\breve{\mathbf{e}}^{m,n (k)}_{i} &={f_{rel}}(\mathbf{h^\prime}^{m(k)}_{i},\mathbf{h}^{n(k)}_{i}) = Conv(\mathrm{g} (\mathbf{h^\prime}^{m(k)}_{i},\mathbf{h}^{n(k)}_{i}))\in \mathbb{R}^{H\times W \times C},
\label{eq5}
\end{aligned}
\end{equation}
\noindent where $\breve{\mathbf{e}}^{m,n}_{i} \in \mathbb{R}^{H\times W \times C}$ is a 3D tensor, which contains the hidden representation of the cross-domain relation. The detailed architecture can be found in Fig.~\ref{fig:3}.  

\subsubsection{Cross-scale Message Aggregation.} 

In our {\scshape{HyGnn}}, we employ different aggregation schemes for each node to aggregate feature messages from its neighbors. For the message $\mathbf{m}^m_{i,j}$ passed from node $v^m_i$ to $v^m_j$ within the same domain across different scales, we have:
\begin{equation}
\begin{aligned}
\mathbf{m}^{m(k)}_{i,j}= M(\mathbf{h}^{m(k-1)}_{i}, \mathbf{e}_{i,j}^{m(k-1)})=Sig(\mathbf{e}_{i,j}^{m(k-1)})\cdot \mathbf{h}^{m(k-1)}_{i},
\label{eq6}
\end{aligned}
\end{equation}
\noindent where $M(\cdot)$ is the cross-scale message passing function (aggregator), and $Sig(\cdot)$ maps the edge's embedding into the link weight. Note that since our {\scshape{HyGnn}} is devised to handle the pixel-level task, the link weight between nodes is in the manner of a 2D map. Thus, $\mathbf{m}^m_{i,j}$ assigns the pixel-wise weighted features from node $v^m_i$ to $v^m_j$ to aggregate information.

\subsubsection{Cross-domain Message Aggregation.} 

As the cross-domain discrepancy is significant in the high-dimensional feature space and distribution, directly passing the learned representations of one node to its neighboring nodes for aggregation is a sub-optimal solution. Therefore, we formulate the message passing from node $v^m_i$ to $v^n_i$ as an adaptive representation learning process conditioned on $\mathbf{h}^{n}_{i}$. Here, we use the similar idea with that used in the cross-domain edge embedding process, \emph{i.e.,} using a one-shot adapter to predict the message that should be passed: 
\begin{equation}
\begin{aligned}
\breve{\mathbf{m}}^{m,n(k)}_{i}&= \breve{M}(\mathbf{h}^{m(k-1)}_{i}, \breve{\mathbf{e}}^{m,n (k-1)}_{i} {\big|\big|} \mathbf{h}^{n(k-1)}_{i})\\
&=\breve{\mathcal{A}}_m(Sig(\breve{\mathbf{e}}^{m,n (k-1)}_{i})\cdot \mathbf{h}^{m(k-1)}_{i} {\big|\big|} \mathbf{h}^{n(k-1)}_{i})\\
&=\breve{E}_\eta(Sig(\breve{\mathbf{e}}^{m,n (k-1)}_{i})\cdot \mathbf{h}^{m(k-1)}_{i}) * \mathbf{h}^{n(k-1)}_{i}\\
&={\psi^{*}}_i^{m(k-1)} * \mathbf{h}^{n(k-1)}_{i},
\label{eq7}
\end{aligned}
\end{equation}
\noindent where $\breve{M}(\cdot$) means the message passing function between nodes from two different domains. $\breve{\mathcal{A}}(\cdot)$ is the adapter which is conditioned on the node embedding of target domain $\mathbf{h}^{n(k-1)}_{i}$. $\breve{E}_\eta(\cdot)$ means a small CNN with learnable parameters $\eta$, which serves as an one-shot learner to predict the dynamic parameters. ${\psi^{*}}_i^{m(k-1)}$ is the produced dynamic convolutional kernels, which includes the guidance information that should be propagated from node $v^m_i$ to $v^n_i$.

\subsubsection{Two-stage Node State Update.} In the $k_{th}$ step, our {\scshape{HyGnn}} first aggregates the information from the cross-domain nodes within the same scale $i$ using Eq.~\ref{eq7}. Therefore, $v^n_i$ ($i \in \{1,\cdots,N\}$ \& $n \in \{1,2\}$) gets an intermediate state $\widetilde{\mathbf{h}}^{n(k)}_i$ by taking into account its received cross-domain message $\breve{\mathbf{m}}^{m,n(k)}_{i}$ and its prior state $\mathbf{h}^{n(k-1)}_i$. Here, following~\cite{qi2018learning}, we apply Gated Recurrent Unit (GRU)~\cite{ballas2015delving} as the update function,
\begin{equation}
\begin{aligned}
\widetilde{\mathbf{h}}^{n(k)}_i= U_{GRU} (\mathbf{h}^{n(k-1)}_i, \breve{\mathbf{m}}^{m,n(k)}_{i}).
\label{eq8}
\end{aligned}
\end{equation}
Then, {\scshape{HyGnn}} performs message passing across scales within the same domain $n$ using Eq.~\ref{eq3}, and aggregates messages using Eq.~\ref{eq6}. After that, $v^n_i$ gets the new state $\mathbf{h}^{n(k)}_i$ after the $k_{th}$ iteration by considering the cross-scale message $\mathbf{m}^{n(k)}_{j,i}$ and its intermediate state $\widetilde{\mathbf{h}}^{n(k)}_i$,
\begin{equation}
\begin{aligned}
\mathbf{h}^{n(k)}_i = U_{GRU} (\widetilde{\mathbf{h}}^{n(k)}_i, \mathbf{m}^{n(k)}_{j,i}).
\label{eq9}
\end{aligned}
\end{equation}

\subsubsection{Readout Function.} After K message passing iterations, the updated multi-scale features of two domains $\mathcal{H}_1=\{\mathbf{h}^{1}_{i}\}_{i=1}^{|\mathcal{V}^1|}$ and $\mathcal{H}_2=\{\mathbf{h}^{2}_{i}\}_{i=1}^{|\mathcal{V}^2|}$ are merged to form their final representations $\mathbf{f}_{d}$ and $\mathbf{f}_{l}$, 
\begin{equation}
\begin{aligned}
\mathbf{f}_{d} = \mathcal{C}_d(\mathcal{H}_1) ~~and~~ \mathbf{f}_{l} = \mathcal{C}_l(\mathcal{H}_2), 
\label{eq10}
\end{aligned}
\end{equation}
where $\mathcal{C}_d(\cdot)$ and $\mathcal{C}_l(\cdot)$ are the merge functions by concatenation. Then, $\mathbf{f}_{d}$ and $\mathbf{f}_{l}$ are fed into a convolution layer to get the final per-pixel predictions.

\subsubsection{Loss.} Our {\scshape{HyGnn}} is implemented to be fully differentiable and end-to-end trainable. The loss for each task can be computed after the readout functions, and the error can propagate back according to the chain rule. Here, we simply employ the Mean Square Error (MSE) loss to optimize the network parameters for two tasks:
\begin{equation}
\begin{aligned}
L= L_1(\mathcal{D}^g,  \mathcal{D}) + \lambda L_2(\mathcal{L}^g,  \mathcal{L}),
\end{aligned}
\end{equation}

\noindent where $L_1$ and $L_2$ are MSE losses, and $\lambda$ is the combination weight. As our main task is the crowd counting, we set $\lambda = 0.001$ to emphasize the accuracy of counting results. 

\begin{table*}[pt]
	\centering
	\caption{Comparison with other state-of-the-art crowd counting methods on four benchmark crowd counting datasets using the MAE and MSE metrics.}\label{tab2}
	\resizebox{0.95\textwidth}{!}{
		\begin{tabular}{l|cc|cc|cc|cc}
			\hline
			\multirow{2}{*}{\quad Methods} &\multicolumn{2}{c|}{Shanghai Tech A} &\multicolumn{2}{c|}{Shanghai Tech B} &\multicolumn{2}{c|}{UCF\_CC\_50} &\multicolumn{2}{c}{UCF\_QNRF} \\
			\cline{2-9}
			&MAE &MSE &MAE &MSE &MAE &MSE &MAE &MSE\\
			
			\hline
			
			\quad{Crowd CNN~\cite{zhang2015cross}} &181.8 &277.7 &32 &49.8 &467 &498 &- &- \\

			\quad{MC-CNN~\cite{zhang2016single}} &110.2 &173.2 &26.4 &41.3 &377.6 &509.1 &277 &426 \\
			
			\quad{Switching CNN~\cite{sam2017switching}} &90.4 &135 &21.6 &33.4 &318.1 &439.2 &228 &445\\

			\quad{CP-CNN~\cite{sindagi2017generating}} &73.6 &106.4 &20.1 &30.1 &298.8 &320.9 &- &-\\
			
			\quad{D-ConvNet~\cite{shi2018crowd}} &73.5 &112.3 &18.7 &26 &288.4 &404.7 &- &- \\

			\quad{L2R~\cite{liu2018leveraging}} &72 &106.6 &13.7 &21.4 &279.6 &388.9 &- &- \\

			\quad{CSRNet~\cite{li2018csrnet}} &68.2 &115 &10.6 &16 &266.1 &397.5 &- &-\\

			\quad{PACNN~\cite{shi2019revisiting}} &66.3 &106.4 &8.9 &13.5&267.9 &357.8 &- &-\\
			\quad{RA2-Net~\cite{Liu_2019_CVPR}} &65.1 &106.7 &8.4 &14.1 &- &- &116 &195\\
			
			\quad{SFCN~\cite{wang2019learning}} &64.8 &107.5 &7.6 &13 &214.2 &318.2 &124.7 &203.5\\
			
			\quad{TEDNet~\cite{Jiang_2019_CVPR}} &64.2 &109.1 &8.2 &12.8 &249.4 &354.2 &113 &188\\

			\quad{ADCrowdNet~\cite{liu2019adcrowdnet}} &{63.2} &{98.9} &7.6 &13.9 &257.1 &363.5 &- &-\\
			
			\quad{{\scshape{HyGnn}} (Ours)} &{\bfseries 60.2} &{\bfseries 94.5} &{\bfseries 7.5} &{\bfseries 12.7} &{\bfseries 184.4} &{\bfseries 270.1} &{\bfseries 100.8} &{\bfseries 185.3}{\tiny }\\

			\hline
		\end{tabular}
	}
	\label{tab:2}
\end{table*}
\section{Experiments} 

In this section, we empirically validate our {\scshape{HyGnn}} on four public counting benchmarks (\emph{i.e.,} ShanghaiTech Part A, ShanghaiTech Part B, UCF\_CC\_50 and UCF\_QNRF). First, we conduct an ablation experiment to prove the effectiveness of our hybrid graph model and the multi-task learning. Then, our proposed {\scshape{HyGnn}} is evaluated on all of these public benchmarks, and compare the performance with the state-of-the-art approaches. 

\subsubsection{Datasets.}
 We use Shanghai Tech~\cite{zhang2016single}, UCF\_CC\_50~\cite{idrees2013multi} and UCF\_QNRF~\cite{idrees2018composition} for benchmarking our {\scshape{HyGnn}}. Shanghai Tech provides 1,198 annotated images containing more than 330K people with head center annotations. It includes two subsets: Shanghai Tech A and Shanghai Tech B. UCF\_CC\_50 provides 50 images with 63,974 head annotations in total. The small dataset volume and large count variance make it a very challenging dataset. UCF\_QNRF is the largest dataset to date, which contains 1,535 images that are divided into training and testing sets of 1,201 and 3,34 images respectively. All of these benchmarks have been widely used for performance evaluation by state-of-the-art approaches. 

\subsubsection{Implementation Details and Evaluation Protocol.}
To make a fair comparison with existing works, we use a truncated VGG as the backbone network. Specifically, the first 10 convolutional layers from VGG-16 are used as the front-end and shared by two tasks. Following~\cite{li2018csrnet}, our counting and localization back-ends are composed of 8 dilated convolutions with kernel size $3\times3$.

We use Adam optimizer with an initial learning rate $10^{-4}$. We set the momentum to $0.9$, the weight decay to $10^{-4}$ and the batchsize to $8$. For data augmentation, the training images and corresponding groundtruths are randomly flipped and cropped from different locations with the size of $400 \times 400$. In the testing phase, we simply feed the whole image into the model for predicting the counting and localization results. 

We adopt Mean Absolute Error ($\rm {MAE}$) and Mean Squared Error ($\rm {MSE}$) to evaluate the performance. The definitions are as follows:
\begin{equation}\small
\begin{aligned}
\rm {MAE} = \frac{1} {N} \sum_{i=1}^{N} |C_i - C_i^{GT}| ~~and~~ \rm {MSE} = \sqrt{\frac{1} {N} \sum_{i=1}^{N} |C_i - C_i^{GT}|^2}, 
\label{eq11}
\end{aligned}
\end{equation}
where $C_i$ and $C_i^{GT}$ are the estimated count and the ground truth of the $i_{th}$ testing image, respectively.

\begin{table}[pt]
	\centering
	\caption{Analysis of the proposed method. Our results are obtained on Shanghai Tech A.} \label{tab1}
	\resizebox{0.45\textwidth}{!}{
		\begin{tabular}{l|cc}
			
			\quad Methods &MAE$\downarrow$ &MSE$\downarrow$\\
			\hline
			
			\quad{Baseline Model (a truncated VGG)} &68.2&115.0\\
			\quad{Baseline + PSP~\cite{zhao2017pyramid}} &65.3&106.8\\
			\quad{Baseline + Bidirectional Fusion~\cite{yang2018multi}} &65.1&105.9\\
			\quad{Single-task GNN} &62.5&103.4\\
			\quad{Multi-task GNN \emph{w/o} adapter} &62.4&101.8\\
			\hline
			\quad{{\scshape{HyGnn}} (N=2, K=3)} &62.1&100.8\\	
			\quad{{\scshape{HyGnn}} (N=3, K=3)} &60.2&94.5\\
			\quad{{\scshape{HyGnn}} (N=5, K=3)} &60.2&94.1\\
			\hline
			\quad{{\scshape{HyGnn}} (N=3, K=1)} &65.4&109.2\\	
			\quad{{\scshape{HyGnn}} (N=3, K=3)} &60.2&94.5\\
			\quad{{\scshape{HyGnn}} (N=3, K=5)} &60.1&94.4\\
			\hline				
			
			\quad{\bf{\scshape{HyGnn}} (full model)} &{\bfseries 60.2}&{\bfseries 94.5}\\
			
		\end{tabular}
	}
	\label{tab:1}
\end{table}

\subsubsection{Ablation Study.}
Extensive ablation experiments are performed on ShanghaiTech A to verify the impact of each component of our {\scshape{HyGnn}}. Results are summarized in Tab.~\ref{tab:1}. 

\noindent {\bfseries \small{Effectiveness of {\scshape{HyGnn}}.}}
To show the importance of our {\scshape{HyGnn}}, we offer a baseline model without {\scshape{HyGnn}}, which gives the results from our backbone model, the truncated VGG with dilated back-ends. As shown in Tab.~\ref{tab1}, our {\scshape{HyGnn}} significantly outperforms the baseline by 8.0 in MAE  ($68.2\mapsto 60.2$) and 20.5 in MSE ($115.0\mapsto 94.5$). This is because our {\scshape{HyGnn}} can simultaneously model the multi-scale and cross-domain relationships which are important for achieving accurate crowd counting results.

\noindent {\bfseries \small{Multi-task GNN vs. Single-task GNN.}}
To evaluate the advantage of multi-task cooperation, we provide a single-task model which only formulates the cross-scale relationship. According to our experiments, {\scshape{HyGnn}} outperforms the single-task graph neural network by 2.3 in MAE ($62.5\mapsto 60.2$)  and 8.9 in MSE ($103.4\mapsto 94.5$). This is because our {\scshape{HyGnn}} is able to distill mutual benefits between the density and localization, while single-task graph neural network ignores these important information. 

\noindent {\bfseries \small{Effectiveness of the Cross-domain Edge Embedding.}} Our {\scshape{HyGnn}} carefully deals with the cross-domain information by a learnable adapter. To evaluate its effectiveness, we provide a multi-task GNN without the learnable adapter. Instead, we directly fuse features from different domains through the aggregation operation. As shown Tab.~\ref{tab1}, our cross-domain edge embedding method achieves better performance in both MAE ($60.2$ vs. $62.4$) and MSE ($94.5$ vs. $101.8$), which indicates that our design of cross-domain edge embedding method is helpful for better leveraging the information from the other domain.

\begin{figure*}[pt]
	\begin{center}
		\includegraphics[width=0.95\linewidth]{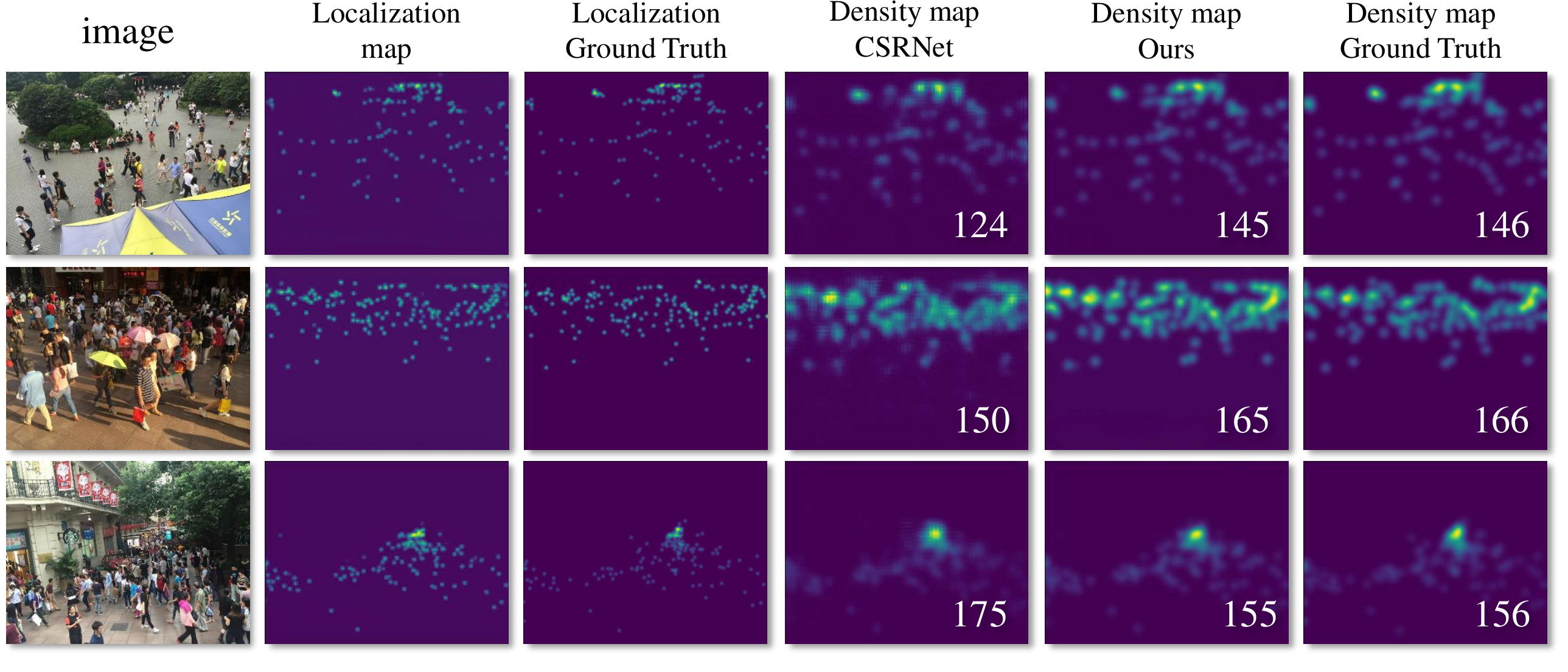}
	\end{center}
	\caption{Density and localization maps generated by our {\scshape{HyGnn}}. We also show the counting map estimated by CSRNet for comparison. Clearly, our {\scshape{HyGnn}} produces more accurate results.}
	\label{fig:vis}
\end{figure*}

\noindent {\bfseries \small{Node Numbers $N$ in {\scshape{HyGnn}}.}} In our model, we have $N$ numbers of nodes in each domain, \emph{i.e.}, $|\mathcal{V}^1| = |\mathcal{V}^2| = N$. To investigate the impact of node numbers, we report the performance of our {\scshape{HyGnn}} with different $N$. We find that with more scales in the model ($2\mapsto3$), the performance improves significantly (\emph{i.e.,} $62.1\mapsto 60.2$ in MAE and $100.8\mapsto 94.5$ in MSE). However, when further considering more scales ($3\mapsto5$), it only achieves slight performance improvements, \emph{i.e.,} $60.1\mapsto 60.2$ in MAE and $94.5\mapsto 94.1$ in MSE. This may be due to the redundant information within additional features. Considering the tradeoff between efficiency and performance, we set $N=3$ in the following experiments. 


\noindent {\bfseries \small{Message Passing Iterations $K$.}} To evaluate the impact of message passing iterations $K$, we report the performance of our model with different passing iterations. Each message passing iteration in {\scshape{HyGnn}} includes two cascade steps: i) the cross-scale message passing and ii) the cross-domain message passing. We find that with more iterations ($1\mapsto3$), the performance of our model improves to some extent. When further considering more iterations  ($3\mapsto5$), it just bring a slight improvement. Therefore, we set $k=3$, and our {\scshape{HyGnn}} can converge to an optimal result.

\noindent {\bfseries \small{GNN vs. Other Multi-feature Aggregation Methods.}}
Here, we conduct an ablation to evaluate the superiority of GNN. To prevent other jamming factor, we use a single-task GNN to fully distill the underlying relationships between multi-scale features, and compare our method with two well-known multi-scale feature aggregation methods (PSP~\cite{zhao2017pyramid} and Bidirectional Fusion~\cite{yang2018multi}). As can be seen, our GNN-based method greatly outperforms other methods by a large margin.

\subsubsection{Comparison with State-of-the-art.} 
We compare our {\scshape{HyGnn}} with the state-of-the-art for the performance of counting. 

\noindent {\bfseries \small{Quantitative Results.}} As can be seen in Tab.~\ref{tab:2}, our {\scshape{HyGnn}} consistently achieves better results than other methods on four widely-used benchmarks. Specifically, our method greatly outperforms previous best result by $3.0$ in MAE and $4.4$ in MSE on ShanghaiTech Part A. Although previous methods have made remarkable progresses on ShanghaiTech Part B, our {\scshape{HyGnn}} also achieves the best performance. Compared with existing top approaches like ADCrowdNet~\cite{liu2019adcrowdnet} and SFCN~\cite{wang2019learning}, our {\scshape{HyGnn}} achieves performance gain by $0.1$ in MAE and $1.2$ in MSE and $0.1$ in MAE and $0.3$ in MSE, respectively. On the most challenging UCF\_CC\_50, our {\scshape{HyGnn}} achieves considerable performance gain by decreasing the MAE from previous best $214.2$ to $184.4$ and MSE from $318.2$ to $270.1$. On UCF-QNRF dataset, our {\scshape{HyGnn}} also outperforms other methods by a large margin. As shown in Tab.~\ref{tab:2}, our {\scshape{HyGnn}} achieves a significant improvement of $10.8\%$ in MAE over the existing best result produced by TEDNet~\cite{Jiang_2019_CVPR}. Compared with other top-ranked methods, our {\scshape{HyGnn}} produces more accurate results. This is because {\scshape{HyGnn}} is able to leverage \emph {free-of-cost} localization information and jointly reason all relations among them. 

\noindent {\bfseries \small{Qualitative Results.}} Fig.~\ref{fig:vis} provides some visualization comparisons of the predicted density maps and counts with CSRNet~\cite{li2018csrnet}. In addition, we also show the localization results. We observe that our {\scshape{HyGnn}} is very powerful, achieves much more accurate count estimations and reserves more consistency with the real crowd distributions. This is because our {\scshape{HyGnn}} can distill the significant benefit information from the auxiliary task through a graph. 


\section{Conclusions} 

In this paper, we propose a novel method for crowd counting with a hybrid graph model. To best of our knowledge, it is the first deep neural network model that can distill both multi-scale and mutual beneficial relations within a unified graph for crowd counting. The whole {\scshape{HyGnn}} is end-to-end differentiable, and is able to handle different relations effectively. Meanwhile, the domain gap between different tasks is also carefully considered in our {\scshape{HyGnn}}. According to our experiments, {\scshape{HyGnn}} achieves significant improvements compared to recent state-of-the-art methods on four benchmarks. We believe that our {\scshape{HyGnn}} can also incorporate other knowledge, \emph{e.g.,} foreground information, for further improvements.

\noindent{\bf Acknowledgement.} This work was supported in part by the National Key R\&D Program of China (No.2017YFB1302300) and the NSFC (No.U1613223).

{\small
	\bibliographystyle{aaai}
	\bibliography{aaai}
}

\end{document}